\documentclass{article}
\usepackage{spconf,amsmath,graphicx,hyperref}
\usepackage[utf8]{inputenc} 
\usepackage[T1]{fontenc}    
\usepackage{hyperref}       
\usepackage{url}            
\usepackage{booktabs}       
\usepackage{amsfonts}       
\usepackage{nicefrac}       
\usepackage{microtype}      
\usepackage{xcolor}         
\usepackage{graphicx}
\usepackage{amsfonts}       
\usepackage{enumitem}       
\usepackage{subcaption}
\usepackage{caption}
\usepackage{wrapfig}
\usepackage{lipsum}
\usepackage{amsmath}
\usepackage{amssymb}

\usepackage[table,xcdraw]{xcolor}
\usepackage{multirow}
\usepackage{array}

\title{Improving Robustness in Sparse Autoencoders via Masked Regularization}
\name{Vivek Narayanaswamy\textsuperscript{*}, Kowshik Thopalli\textsuperscript{*}, Bhavya Kailkhura, Wesam Sakla}
\address{Lawrence Livermore National Laboratory, CA, USA}

\begin{document}

\maketitle
\begin{abstract}

Sparse autoencoders (SAEs) are widely used in mechanistic interpretability to project LLM activations onto sparse latent spaces. However, sparsity alone is an imperfect proxy for interpretability, and current training objectives often result in brittle latent representations. SAEs are known to be prone to feature absorption, where general features are subsumed by more specific ones due to co-occurrence, degrading interpretability despite high reconstruction fidelity. Recent negative results on Out-of-Distribution (OOD) performance further underscore broader robustness related failures tied to under-specified training objectives. We address this by proposing a masking-based regularization that randomly replaces tokens during training to disrupt co-occurrence patterns. This improves robustness across SAE architectures and sparsity levels reducing absorption, enhancing probing performance, and narrowing the OOD gap. Our results point toward a practical path for more reliable interpretability tools.

\end{abstract}
\renewcommand\thefootnote{}
\footnotetext{*equal contribution. This work was performed under the auspices of the U.S. Department of Energy by the Lawrence Livermore National Laboratory under Contract No. DE-AC52-07NA27344, Lawrence Livermore National Security, LLC. and was supported by the LLNL-LDRD Program under Project No. 25-SI-001.LLNL-CONF-2011550
}
\addtocounter{footnote}{-1}
\section{Introduction}

Sparse autoencoders (SAEs) have emerged as key tools in mechanistic interpretability (MI), enabling human-interpretable explanations of large language model (LLM) internals. They do so by mapping dense activations from LLMs into sparse, overcomplete latent representations that reveal underlying structure~\cite{cunningham2023sparse, gao2024scaling, rajamanoharan2024improving, bussmann2025learning, inaba2025how}. The use of SAEs for MI is motivated by the superposition principle~\cite{elhage2022toy, sharkey2022taking}, which posits that individual neurons encode polysemantic mixtures of features, hindering direct interpretation. By enforcing sparsity, SAEs aim to disentangle these features into monosemantic components, enabling human-interpretable analysis of model behavior. However, recent studies~\cite{rajamanoharan2024improving, paulo2025transcoders, rajamanoharan2024jumping} demonstrate that sparsity is an imperfect proxy for interpretability,as enforcing excessive sparsity often biases SAEs toward representations that obscure the structure (e.g., hierarchy) of real-world features.


One of the key problems stemming from the mismatch between sparsity objectives and the hierarchical structure of real-world features is \emph{feature absorption}~\cite{chaninabsorption, bussmann2025learning}. For e.g., a latent meant to represent “words starting with S” may collapse into one representation for “short words starting with S,” underrepresenting the broader concept. While reconstruction remains accurate with fewer active latents, the learned features become harder to interpret. This stems from the SAE's tendency to create shortcuts when words/tokens frequently co-occur—favoring latents that absorb general concepts into more specific ones to satisfy sparsity. These shortcuts hinder interpretability as they fragment general features into incomplete or overly specialized ones thus producing sub-optimal representations. Moreover, recent negative results on the poor OOD generalization performance of probes trained on SAE latents~\cite{kantamneni2025sparse, smith2025negative} demonstrate that SAEs produce brittle representations under distribution shifts. Although absorption and OOD fragility manifest differently, we posit that both arise from inadequately constrained training objectives that fail to prevent shortcut-based representations in current SAEs.



To address this, we introduce a simple yet effective regularization mechanism that mitigates feature absorption by disrupting co-occurrence patterns in text. Specifically, during training, we randomly replace tokens in the input sequence with a fixed mask string (e.g., "...") at a user-defined probability. We observe that this strategy breaks spurious correlations and encourages the SAE to learn more generalizable structure, reducing its reliance on shortcuts. When applied across multiple LLMs (Pythia-160M-deduped, Gemma-2-2B), this strategy consistently reduces absorption and interestingly improves performance on a suite of evaluation metrics~\cite{karvonen2025saebench}. Encouragingly, it also enhances OOD performance~\cite{kantamneni2025sparse}, narrowing the gap with oracle probes. Overall, our results demonstrate that this strategy improves SAE robustness paving the path for more reliable and interpretable tools.



\section{Approach}

\noindent\textbf{Preliminaries}. Let $\mathcal{G}$ denote an LLM operating on a text sequence $\mathbf{t} = [t_1, t_2, \ldots, t_n]$ which are then tokenized. For a given layer $l$, the hidden activations are denoted as $\mathbf{X}^{(l)} = [\mathbf{x}_1^{(l)}, \mathbf{x}_2^{(l)}, \ldots, \mathbf{x}_n^{(l)}]$, where $\mathbf{x}_i^{(l)} \in \mathbb{R}^D$ and $D$ is the activation dimension. These token-level activations serve as training data for the SAE. Let $f$ denote the SAE, which consists of an encoder $e$ that maps token activations into a sparse latent representation $\mathbf{e}(\mathbf{x})$, and a decoder $d$ (or dictionary~\cite{bricken2023towards}) that reconstructs that activation. Specifically, the encoder is defined as $e(\mathbf{x}) = \sigma(W_{\text{e}} \mathbf{x} + \mathbf{b}_{\text{e}}) \in \mathbb{R}^m$ with $m \gg D$, where $W_{\text{e}} \in \mathbb{R}^{m \times D}$, $\mathbf{b}_{\text{e}} \in \mathbb{R}^m$, and $\sigma(\cdot)$ is a sparsity-inducing nonlinearity (e.g., BatchTopK~\cite{bussmann2024batchtopk}). The decoder reconstructs the activation as $\hat{\mathbf{x}} = W_{\text{d}} e(\mathbf{x}) + \mathbf{b}_{\text{d}}$, where $W_{\text{d}} \in \mathbb{R}^{D \times m}$ and $\mathbf{b}_{\text{d}} \in \mathbb{R}^D$. The SAE training objective balances reconstruction fidelity with latent sparsity. Given input activations $\mathbf{x}$ and reconstructed output $\hat{\mathbf{x}}$, the SAE training objective is defined as $\mathcal{L}(\mathbf{x}) = \| \mathbf{x} - \hat{\mathbf{x}} \|_2^2 + \lambda \| e(\mathbf{x}) \|_1$, where $\lambda$ controls the reconstruction and sparsity trade-off. While sparsity is often encouraged via $\ell_1$ regularization, practical implementations commonly apply hard constraints such as Top-$K$~\cite{gao2024scaling} or BatchTop-$K$~\cite{bussmann2024batchtopk} selection over $e(\mathbf{x})$ to limit active latents.

\noindent\textbf{Motivation}. SAE training involves a fundamental trade-off: minimizing reconstruction favors dense representations, while enforcing sparsity encourages fewer active latents. This tension often yields brittle solutions that satisfy the objective but fail to capture semantically coherent structure. As a result, hierarchical or overlapping features are under-represented, and shortcut latents frequently emerge under co-occurrence. Because real-world features are inherently hierarchical, imposing sparsity independently across latents misaligns with the true feature space. These shortcomings manifest as feature absorption and poor OOD performance, both symptomatic of under-specified training objectives.  
Recent architectural advances, such as the \texttt{MatryoshkaBatchTopK} SAE~\cite{bussmann2025learning}, build on Matryoshka representation learning~\cite{kusupati2022matryoshka} to construct nested encoders operating at multiple scales, achieving notable progress toward mitigating these issues. However, as our results show, substantial gaps remain between the OOD generalization of probes trained on SAE activations and oracle probes trained directly on raw LLM activations, along with continued susceptibility to feature absorption.  

We argue that these challenges cannot be overcome by architectural modifications alone, but require stronger training objectives. We posit that combining architectural advances with improved objectives can substantially mitigate shortcut learning in SAEs. To this end, we introduce a simple, architecture-agnostic regularization strategy that suppresses shortcuts and encourages robust, transferable features.


\noindent\textbf{Masking Based Regularization}. For a given input sequence $\mathbf{t}$, we sample a binary mask $\omega \sim \text{Bernoulli}(p)$, where $p$ is a user-defined masking probability. We replace the selected tokens with a special token ``\ldots'' before feeding the sequence into the LLM:
\[
\mathbf{t}' = [t_1', \ldots, t_n'], \quad t_i' =
\begin{cases}
\texttt{‘...’}, & \omega_i = 1 \\
t_i, & \omega_i = 0.
\end{cases}
\]
The LLM activations of the modified tokens are used as input to the SAE. The training objective remains the same but is now applied over masked activations: $\mathcal{L}_{\text{mask}}(\mathbf{x}') = \| \mathbf{x}' - \hat{\mathbf{x}}' \|_2^2 + \lambda \| e(\mathbf{x}') \|_1$. The key rationale is that introducing masking alters the contextual embeddings of surrounding tokens, thereby decorrelating feature co-occurrence and discouraging the SAE from collapsing broad features into overspecialized ones. This forces the latents to capture more generalizable structure, rather than favoring shortcuts, lowering the risk of feature absorption.

\begin{table*}[htbp]
\caption{Performance comparison of the proposed masking strategy on the \texttt{MatryoshkaBatchTopK} SAE trained on Layer 8 of Pythia-160m-deduped across multiple evaluation metrics and sparsity levels. Across metrics, we see consistent improvements with our approach with a small trade-off in explained variance at lower sparsities. Overall, the method demonstrates robustness and strong performance across sparsity regimes.}
\centering
\renewcommand{\arraystretch}{1.2}
\setlength{\tabcolsep}{8pt}
\resizebox{0.8\textwidth}{!}{
\begin{tabular}{
    >{\centering\arraybackslash}p{3.4cm}|
    >{\centering\arraybackslash}p{2.2cm}|
    >{\centering\arraybackslash}m{1.5cm}|
    >{\centering\arraybackslash}m{1.5cm}|
    >{\centering\arraybackslash}m{1.5cm}|
    >{\centering\arraybackslash}m{1.5cm}
}
\rowcolor{gray!30}
\hline
\textbf{Metric} & \textbf{Method} & $\boldsymbol{\ell_0=20}$ & $\boldsymbol{\ell_0=40}$ & $\boldsymbol{\ell_0=80}$ & $\boldsymbol{\ell_0=160}$ \\
\hline

\multirow{2}{*}{Mean Full Absorption ($\uparrow$)} 
& w/o Masking & 86.119 & 91.646 & 94.650 & 97.434 \\
& w/ Masking & \textbf{88.475} & \textbf{93.450} & \textbf{96.437} & \textbf{98.003} \\
\hline

\multirow{2}{*}{Explained Variance ($\uparrow$)} 
& w/o Masking & \textbf{72.172} & \textbf{77.421} & 82.303 & 87.841 \\
& w/ Masking & 71.266 & 76.874 & \textbf{82.419} & \textbf{88.823} \\
\hline

\multirow{2}{*}{Sparse Probing ($\uparrow$)} 
& w/o Masking & 73.483 & 75.099 & \textbf{78.705} & 79.574 \\
& w/ Masking & \textbf{75.574} & \textbf{77.112} & 77.749 & \textbf{79.639} \\
\hline

\multirow{2}{*}{TPP ($\uparrow$)} 
& w/o Masking & 10.158 & 18.218 & \textbf{27.033} & \textbf{30.235} \\
& w/ Masking & \textbf{12.430} & \textbf{18.968} & 26.488 & 29.815 \\
\hline

\multirow{2}{*}{SCR ($\uparrow$)} 
& w/o Masking & 19.808 & \textbf{25.843} & 20.132 & 8.441 \\
& w/ Masking & \textbf{20.343} & 25.01 & \textbf{27.626} & \textbf{20.240} \\
\hline

\end{tabular}}

\label{tab:pythia-metric-comparison}
\end{table*}

\section{Experimental Setup and Results}

\noindent\textbf{Implementation Details}.  
We conduct all experiments on Pythia-160M-deduped~\cite{biderman2023pythia} and Gemma-2-2B~\cite{team2024gemma}. We train SAEs for a total of 500M tokens on the Pile-CC-deduplicated dataset~\cite{gao2020pile}.  To ensure fairness, we adopt the same training setup (hyper-parameters such as batch size, learning rate, etc.) provided in the \texttt{dictionary\_learning}\footnote{\url{https://github.com/saprmarks/dictionary_learning}} code base.  We train SAEs with a dictionary size of $4096$ on residual stream activations from layer 8 of the Pythia-160M-deduped model and layer 12 of the Gemma-2-2B~\cite{karvonen2025saebench}. 
For all experiments, we utilize the recently proposed \texttt{MatryoshkaBatchTopK} architecture~\cite{bussmann2025learning} which has been shown to achieve state-of-the-art performance across a variety of interpretability benchmarks~\cite{karvonen2025saebench} across different SAE variants~\cite{bussmann2024batchtopk}. Moreover, based on~\cite{karvonen2025saebench}, we train our SAEs across sparsity levels ($\ell_{0}$) ranging from 20, 40, 80, and 160. We set the default masking probability $p= 0.3$ as we observe that this value yields the best performance across all metrics (Table \ref{tab:masking-sweep-results}).  


\noindent\textbf{Metrics}. 
We employ a comprehensive suite of five evaluation metrics to assess the performance and robustness of SAEs using our proposed objective.  We provide a brief overview of each metric below and refer the reader to~\cite{karvonen2025saebench} for full methodological details and implementation specifics:
(i)~\underline{Mean Full Absorption}: Measures the extent to which one latent consistently activates in the presence of another, indicating redundancy and reduced specificity;
(ii)~\underline{Explained Variance}: Quantifies the proportion of variance in original activations captured by SAE reconstructions;
(iii)~\underline{Sparse Probing}: Assesses linear separability of SAE latents via sparse logistic regression across multiple classification tasks;
(iv)~\underline{Targeted Probe Perturbation (TPP)}: Captures the causal utility of class-specific latents by measuring probe accuracy drop under targeted ablations. It must be noted that all metrics are designed such that higher scores indicate better performance. 
(v)~\underline{Spurious Correlation Removal (SCR)}: Measures the ability to ablate spurious latents while preserving task-relevant features;


\noindent\textbf{Results}.
We present the results of our experiments on Pythia-160M-deduped in Table~\ref{tab:pythia-metric-comparison} and Gemma-2-2B in Table~\ref{tab:gemma-metric-comparison} for all five metrics across different sparsity levels. We observe that training with masking consistently improves performance across all metrics, especially at higher sparsity levels (lower $\ell_{0}$ values). As can be seen from the results, incorporating the proposed masked training objective (a.k.a w/ masking) leads to significant reduction in absorption (as evidenced by increased absorption scores). In the case of Pythia-160M-deduped, we observe an improvement of $2.35\%$ points at highest sparsity level ($\ell_{0}=20$). We also observe an striking improvement of $3.75\%$ points at the same sparsity level for Gemma-2-2B, indicating that the benefits of the proposed training objective are consistent across different model sizes. This indicates that our training objective effectively reduces redundancy and absorption in the latent space, leading to more specific and informative representations.  Interestingly, we also observe that the benefits of masking diminish at lower sparsity levels (high $\ell_{0}$). We hypothesize that this is due to the fact that at lower sparsity, the SAE is already able to learn a more diverse set of features because of the increased flexibility to represent information, and therefore the additional masking does not provide significant benefits. This hypothesis is further corroborated by observing that the baseline (a.k.a w/o masking) themselves achieve high absorption scores as sparsity decreases.

\begin{table*}[htbp]

\caption{Performance comparison of the \texttt{MatryoshkaBatchTopK} SAE trained w/ and w/o masking on layer 12 of Gemma2-2-b. The masking strategy tends to provide steady improvements in mean absorption score and explained variance, with benefits visible across most sparsity levels. For other metrics such as sparse probing, TPP, and SCR, the w/ Masking approach also performs competitively, showing strong results overall with few trade-offs appear in a few settings with increase in $\ell_0$.}
\centering
\renewcommand{\arraystretch}{1.2}
\setlength{\tabcolsep}{8pt}
\resizebox{0.8\textwidth}{!}{
\begin{tabular}{
    >{\centering\arraybackslash}p{3.4cm}|
    >{\centering\arraybackslash}p{2.2cm}|
    >{\centering\arraybackslash}m{1.5cm}|
    >{\centering\arraybackslash}m{1.5cm}|
    >{\centering\arraybackslash}m{1.5cm}|
    >{\centering\arraybackslash}m{1.5cm}
}
\rowcolor{gray!30}
\hline
\textbf{Metric} & \textbf{Method} & $\boldsymbol{\ell_0=20}$ & $\boldsymbol{\ell_0=40}$ & $\boldsymbol{\ell_0=80}$ & $\boldsymbol{\ell_0=160}$ \\
\hline

\multirow{2}{*}{Mean Full Absorption ($\uparrow$)} 
& w/o Masking & 90.805 & 97.365 & \textbf{98.969} & 99.174 \\
& w/ Masking & \textbf{94.559} & \textbf{98.753} & 97.322 & \textbf{98.979} \\
\hline

\multirow{2}{*}{Explained Variance ($\uparrow$)} 
& w/o Masking & \textbf{53.516} & 58.984 & 64.453 & 69.922 \\
& w/ Masking & 53.125 & \textbf{59.375} & \textbf{64.844} & \textbf{71.094} \\
\hline

\multirow{2}{*}{Sparse Probing ($\uparrow$)} 
& w/o Masking & \textbf{74.473} & 73.300 & \textbf{74.659} & \textbf{75.659} \\
& w/ Masking & 74.206 & \textbf{77.341} & 74.243 & 75.363 \\
\hline

\multirow{2}{*}{TPP ($\uparrow$)} 
& w/o Masking & 1.000 & \textbf{3.320} & 7.178 & \textbf{17.550} \\
& w/ Masking & \textbf{1.277} & 2.893 & \textbf{7.295} & 15.988 \\
\hline

\multirow{2}{*}{SCR ($\uparrow$)} 
& w/o Masking & \textbf{22.753} & 22.991 & \textbf{31.270} & 28.468 \\
& w/ Masking & 21.216 & \textbf{30.111} & 30.736 & \textbf{32.775} \\
\hline

\end{tabular}
}
\label{tab:gemma-metric-comparison}
\end{table*}

Next, we observe that the explained variance scores in Table~\ref{tab:pythia-metric-comparison} and \ref{tab:gemma-metric-comparison} show a slight decrease (for Pythia) or comparable performance (Gemma) for the proposed approached compared to the no masking baseline at higher sparsity levels. This indicates that the proposed regularization encourages the SAEs to learn more atomic and distinct features. While we may not capture as much variance present in the original activations required for accurate reconstruction, our results show that our latents are more meaningful and less redundant. Furthermore, we observe that the proposed training objective leads to consistent improvements in sparse probing performance across all sparsity levels. For instance, in the case of Pythia-160M-deduped, we observe an improvement of $2.09\%$ points at $\ell_{0}=20$ and $2.01\%$ points at $\ell_{0}=40$. This indicates that the features learned through the proposed training objective are more discriminative and therefore lead to better performance on downstream tasks. We also make similar observations on the spurious correlation removal and targeted probe perturbation metrics. For instance, in the case of Pythia-160M-deduped, we observe an improvement of $2.27\%$ points at $\ell_{0}=20$ on the TPP metric. Encouragingly, we observe consistent improvements or comparable performance across all sparsity levels on the challenging spurious correlation removal metric. For example, we note an improvement of $7.5\%$ points at $\ell_{0}=80$ and $11.8\%$ points at $\ell_{0}=160$. Similar trends are observed for Gemma-2-2B where the proposed approach maintains comparable or slightly deteriorated performance in few sparsity levels. However, at lower sparsity levels we notice a significant improvement of $4.3\%$ points at $\ell_{0}=160$.  
These results clearly indicate that the features learned through the proposed training objective are more robust and less prone to spurious correlations and therefore lead to better performance on downstream tasks.

\begin{figure}[h]
  \centering
  \includegraphics[width=0.7\columnwidth]{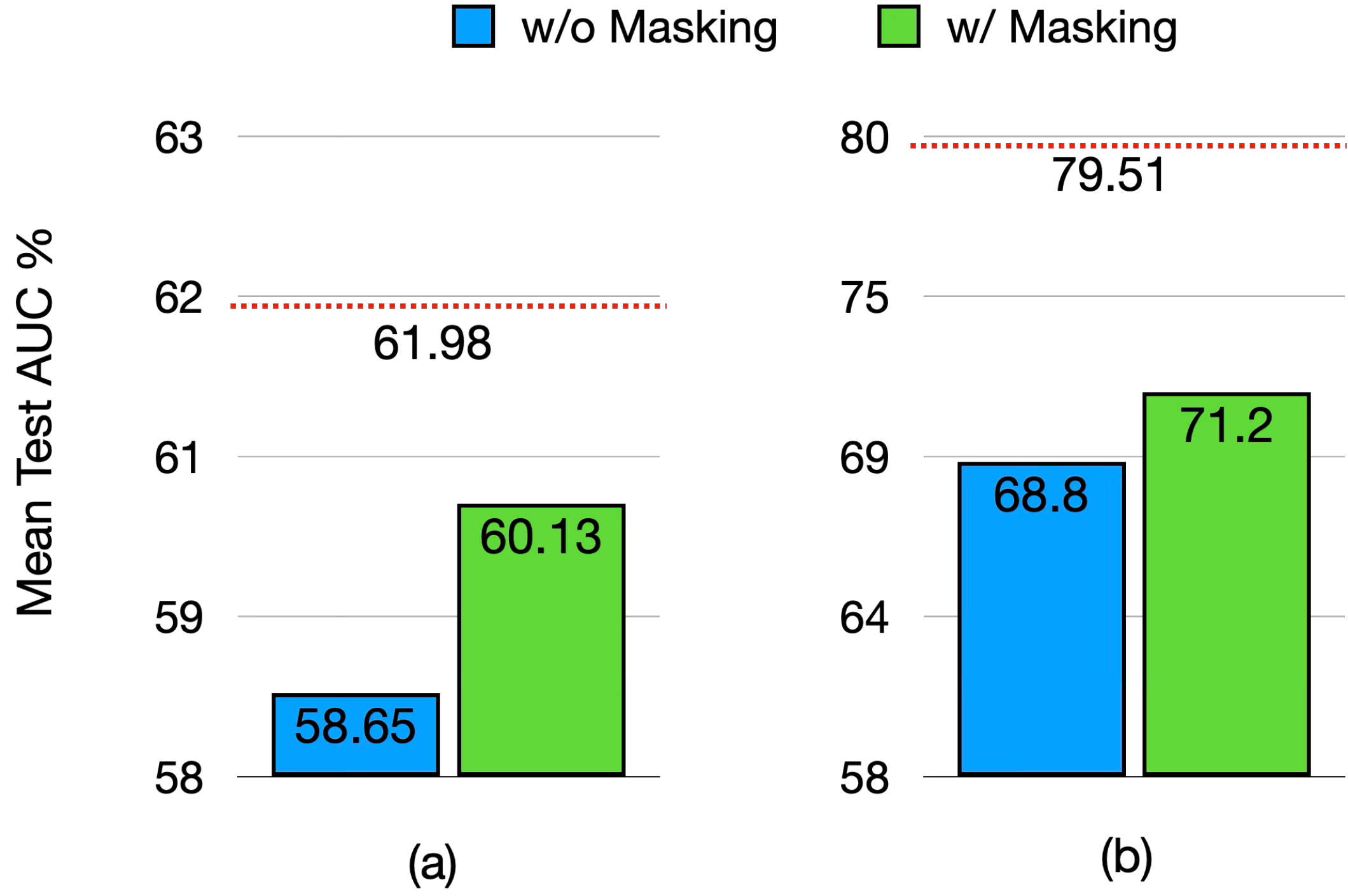} 
  \caption{OOD probing results: Mean test AUC across 8 OOD datasets from~\cite{kantamneni2025sparse} at $\ell_{0}=20$  with \texttt{MatryoshkaBatchTopK} SAEs on Pythia-160m-deduped (a) and Gemma-2-2b (b). While oracle probes (red, trained on raw LLM activations) outperform SAEs, our objective substantially improves OOD performance over the baseline.}
   \label{fig:ood}
\end{figure}

\begin{table*}[htbp]
\caption{Ablation on masking probabilities for Pythia-160m-deduped at $L_0=20$ with \texttt{Matryoshka} SAEs. As a general trend, explained variance decreases as masking increases. We select $m=0.3$ as it achieves the optimal trade-off, while $m=0.0$ corresponds to no masking.}
\centering
\renewcommand{\arraystretch}{1.2}
\setlength{\tabcolsep}{8pt}
\resizebox{0.7\textwidth}{!}{
\begin{tabular}{
    >{\centering\arraybackslash}p{3.6cm}|
    >{\centering\arraybackslash}m{1.5cm}|
    >{\centering\arraybackslash}m{1.5cm}|
    >{\centering\arraybackslash}m{1.5cm}|
    >{\centering\arraybackslash}m{1.5cm}
}
\rowcolor{gray!30}
\hline
\textbf{Metric} & \textbf{$m=0.0$} & \textbf{$m=0.2$} & \textbf{$m=0.3$} & \textbf{$m=0.5$} \\
\hline

Mean Full Absorption ($\uparrow$) & 86.119 & 87.379 & 88.475 & 89.602 \\
\hline
Explained Variance ($\uparrow$)   & 72.171 & 71.860 & 71.270 & 69.540 \\
\hline

\end{tabular}
}
\label{tab:masking-sweep-results}
\end{table*}

\noindent\textbf{Performance on Out-of-Distribution Data}. Recently, Smith \textit{et.al} \cite{smith2025negative} highlighted that SAE probes fail to generalize to OOD tasks and perform worse in comparison to probes trained directly on the LLM activations (oracle). We hypothesize that this poor performance is due to the fact that the current training mechanisms do not encourage the SAEs to learn generalizable features but promotes shortcuts that hinder generalization. Consequently, we also hypothesized that training with masking will also help improve the OOD performance of SAEs.
To validate this, we employ the OOD evaluation protocol provided in~\cite{kantamneni2025sparse}, which involves evaluating on 8 different OOD datasets. We present the mean performance across all datasets for the \texttt{MatryoshkaBatchTopK} SAE at the challenging setting of $\ell_{0}=20$ for both Pythia and Gemma LLMs in Figure~\ref{fig:ood}.
We find that masking leads to significant improvements in OOD performance across both the LLMs ($+1.48\%$ for Pythia and $+2.4\%$ for Gemma) and the gap between the SAE and the oracle is also reduced. These results clearly demonstrate that through such an objective, we can combat the shortcut learning that often plagues SAEs and improve their generalization capabilities.

\noindent\textbf{Ablation on Masking Probability}. 
We conduct an ablation to understand the impact of the masking probability on the performance of the SAE. 
We experiment with masking probabilities in \{ 0.2, 0.3, 0.5\} and present the results in the Table~\ref{tab:masking-sweep-results}. We observe that as masking probability increases, the performance on metrics such as absorption increases however, we observe a tradeoff between reduction in absorption at the cost of reconstruction error measured via explained variance metric. This is likely due to the fact that higher masking probabilities force the model to rely on a smaller subset of features, which may not capture all the relevant information needed for these tasks. Therefore, there exists a trade-off between the atomicity of the learned features (higher with higher masking) and the generalizability of the features to downstream tasks. Consequently, we select $0.3\%$ as our masking probability.


\section{Discussion and Future Work}
\label{sec:conclusion}

We proposed a regularization strategy that mitigates SAE failure modes by breaking co-occurrence patterns during training. Our objective improves performance across metrics, and generalizes across different LLM sizes. It also enhances OOD robustness, a key problem identified with SAEs. We use the mask string \texttt{‘...’} for its neutral role in text, but acknowledge that alternative choices may be more effective in some settings. Unlike token dropout, our approach perturbs context without removing key information. While results are based on Pythia and Gemma, we are extending evaluations to other LLMs of increasing parameter complexity to assess broader generalization. We also plan to apply auto-interpretability techniques~\cite{neuronpedia} to better characterize the learned features.

\vfill\pagebreak

\bibliographystyle{IEEEbib}
\bibliography{main}

\end{document}